# Model-based Deep Medical Imaging: the roadmap of generalizing iterative reconstruction model using deep learning

Jing Cheng[1], Haifeng Wang[1], Yanjie Zhu[1], Qiegen Liu[3], Qiyang Zhang[1], Ting Su[1], Jianwei Chen[1], Yongshuai Ge[1], Zhanli Hu[1], Xin Liu[1], Hairong Zheng[1], Leslie Ying[4], Dong Liang[1, 2]

[1] Paul C. Lauterbur Research Center for Biomedical Imaging, Shenzhen Institutes of Advanced Technology, Chinese Academy of Sciences, Shenzhen, Guangdong, China

[2] Research center for Medical AI, Shenzhen Institutes of Advanced Technology, Chinese Academy of Sciences, Shenzhen, Guangdong, China,  dong.Liang@siat.ac.cn

[3] Department of Electronic Information Engineering, Nanchang University, Nanchang, Jiangxi, China

[4] Departments of Biomedical Engineering and Electrical Engineering, University at Buffalo, the State University of New York, Buffalo, NY 14260 USA

## 1. Introduction

Medical Imaging such as magnetic resonance imaging (MRI), computed tomography (CT), positron emission tomography (PET) is the technique and process of creating visual representations of the interior of a body for clinical analysis and medical intervention, as well as visual representation of the function of some organs or tissues. Currently, medical imaging is playing a more and more important role in clinics. However, there are several issues in different imaging modalities such as slow imaging speed in MRI, radiation injury in CT and PET. Therefore, accelerating MRI, reducing radiation dose in CT and PET have been ongoing research topics since their invention. Usually, acquiring less data is a direct but important strategy to address these issues like undersampling k-space for accelerating MRI, acquiring sparse views in CT and PET imaging. However, less acquisition usually results in aliasing artifacts in reconstructions. Under this circumstance, image reconstruction problem becomes an ill-conditioned problem.

Previous commercial proposals try to mitigate aliasing artifacts based on prior information. In past decades, advanced compressed sensing (CS) uses sparsity prior either in image domain or some transform domains [1-3]. Although this technique has been applied in clinics, there are still some issues to address. For example, it is challenging to determine the numerical uncertainties in the reconstruction model such as the optimal sparse transformations, sparse regularizer in the transform domain, regularization parameters and the parameters of the optimization algorithm. Many attempts have been deliberated to solve these issues, such as some learning-based methods, e.g. dictionary learning [4, 5], and some numerical methods (e.g. L-curve [6], SURE [7]). However, there is no general strategy to overcome these shortcomings.

Recently, deep learning (DL) has been introduced in medical image reconstruction and shown potential on significantly speeding up MR reconstruction and reducing radiation dose [8-20]. In general, DL-based medical imaging can be classified into two categories: data-driven methods and mode-driven methods. The data-driven DL methods adopt an end-to-end fashion to bypass the medical imaging model, by learning the map function between for example undersampled k-space/image and fully-sampled k-space/image in MRI, with the aid of a large training dataset [8-16]. Since its invention in 2016 [8], requiring large size of training dataset and difficult to interpret have been becoming the challenges of data-driven DL methods. Some researchers tried another strategy by starting from an established reconstruction model, and unrolling the procedure of an iterative optimization algorithm to a network, while learning on the different level of variables, such as the regularization parameters, the regularization functions (transformation and regularizer), and even the data consistency metrics (i.e., learning the entire reconstruction model) [17-20]. As a result, such networks can perform well with smaller size of training sets and may be interpretable since it roots from the established image reconstruction model.

However, the widespread application of model-based deep learning methods in medical image reconstruction raises the following questions: 1. given an established CS model and an optimization algorithm, is there a general framework on how to apply deep learning to achieve the best performance? 2. if unrolling all algorithms to their own best performance, will they lead to the same performance? If not, which algorithm is a good starting point? In this paper, we try to answer these questions by proposing a general framework on combining the CS reconstruction model with deep learning, giving the examples to demonstrate the performance and requirements of unrolling different algorithms using deep learning.

## 2. Theory

### 2.1 Medical imaging reconstruction model

Generally, reconstructing medical image $f$ can be formulated as following:

$$f = Am + \delta \quad (1)$$

where $m$ is the vector of pixels we wish to reconstruct from the data $f$, $\delta$ denotes the measurement errors which can be well modeled as noise, $A$ is the encoding matrix. If Nyquist sampling criteria is met, then the image can be obtained directly by decoding e.g. inverse Fourier transform in MRI. Usually, the data is acquired by sub-Nyquist sampling, then, the system matrix $A$ is not well conditioned, and regularization incorporating some prior information is needed to reconstruct medical image. Thus the image reconstruction can be formulated as the following optimization problem:

$$\hat{m} = \underset{m}{argmin} \frac{1}{2} \|Am - f\|_2^2 + \lambda \|\Psi m\|_p \quad (2)$$

$\|\Psi m\|_p$ denotes the regularization term which can enforce prior information that improves image quality for undersampled data. In compressed sensing (CS), the sparsity prior is usually used as the regularization term, where $\Psi$ is the sparse transform and $0 \leq p \leq 1$.

The regularized least-square objective function (2) is the best linear unbiased estimator (BLUE) but may not be the effective one to estimate the image $\hat{m}$ from the partial acquisition $f$. More generally, the reconstruction model can be written as

$$\hat{m} = \underset{m}{argmin}\, F(Am, f) + \lambda R(m) \tag{3}$$

where $F(Am, f)$ denotes data fidelity, $R(m)$ is the regularization function.

**2.2 Model-driven deep learning image reconstruction approaches**

  Model-driven DL methods unroll the iterations of an optimization algorithm for reconstruction problem to a deep network, and learn the constraints and parameters in the reconstruction model from the training data. Thus, the network architectures of these approaches are determined by the data flow of the optimization algorithm. Specifically, starting from a traditional image reconstruction model, a) we first relax the regularizations in the model and unroll the iterations of reconstruction to a learnable deep network architecture. b) then, we further relax the constraints of data consistency in the model, let the network learn the data fidelity freely. c) finally, the fixed structure of variables in the algorithm is broken and the combinations are learned by the network.

  We will take three state-of-the-art CS algorithms as examples to demonstrate the strategy from the specific reconstruction model and optimization algorithm to the learnable architecture step by step. The overall illustration of the algorithms and three learning states is summarized in Table 1.

Table 1. Summarization of the learning states with different algorithms.

| algorithm \ learning state | I | II | III |
|---|---|---|---|
| PDHG-net | $R(m)$, $\|Am - f\|_2^2$ | $R(m)$, $F(Am, f)$ | $R(m), F(Am, f)$, variable combinations |
| ADMM-net | $\|\Psi(m)\|_p$, $\|Am - f\|_2^2$ | $\|\Psi(m)\|_p$, $F(Am, f)$ | $R(m), F(Am, f)$, variable combinations |
| ISTA-net | $\Psi(m)$, $\|Am - f\|_2^2$ | $\Psi(m)$, $F(Am, f)$ | $\Psi(m), F(Am, f)$, variable combinations |

① **the Primal Dual approach**

*PDHG-net-I*

  The primal dual hybrid gradient algorithm, also known as Chambolle-Pock (CP) algorithm, has been applied on several image restoration problems such as de-noising, deconvolution, inpainting, etc [21]. The CP algorithm solves an optimization problem simultaneously with its dual, which provides a robust convergence check – the duality gap. Replacing $\|\Psi m\|_p$ with $R(m)$, using the CP algorithm, the solution of problem (2) is

$$\begin{cases} d_{n+1} = prox_\sigma[F^*](d_n + \sigma A \bar{m}_n) \\ m_{n+1} = prox_\tau[R](m_n - \tau A^T d_{n+1}) \\ \bar{m}_{n+1} = m_{n+1} + \theta(m_{n+1} - m_n) \end{cases} \tag{4}$$

where $\sigma$, $\tau$ and $\theta$ are the algorithm parameters, $F^*$ is the convex conjugate of the function $\frac{1}{2}\|Am - f\|_2^2$, and $prox$ denotes the proximal operator, which can be obtained by the following minimization:

$$prox_\tau[R](x) = \arg\min_z \left\{ R(z) + \frac{\|z-x\|_2^2}{2\tau} \right\} \quad (5)$$

Eq. (4) becomes

$$\begin{cases} d_{n+1} = \frac{d_n + \sigma(A\bar{m}_n - f)}{1+\sigma} \\ m_{n+1} = prox_\tau[R](m_n - \tau A^T d_{n+1}) \\ \bar{m}_{n+1} = m_{n+1} + \theta(m_{n+1} - m_n) \end{cases} \quad (6)$$

Since it is not easy to choose the optimal parameters and transforms, and the simplicity limitation of the convex functions that makes (5) have close-form solution is not always satisfied in practice, a learned operator $\Lambda$ is used to replace the proximal operator $prox_\tau[R]$ and learn the parameters. Thus the algorithm, called PDHG-net-Ⅰ, can be formed as

$$\begin{cases} d_{n+1} = \frac{d_n + \sigma(A\bar{m}_n - f)}{1+\sigma} \\ m_{n+1} = \Lambda(m_n - \tau A^T d_{n+1}) \\ \bar{m}_{n+1} = m_{n+1} + \theta(m_{n+1} - m_n) \end{cases} \quad (7)$$

The parameters $\sigma$, $\tau$ and $\theta$ and the operator $\Lambda$ are all learned by the network. The PDHG-net-Ⅰ learns the regularization functions including transform and regularier through the network implicitly.

### *PDHG-net-II*

If the constraint of data consistency $\|Am - f\|_2^2$ in problem (2) is relaxed as $F(Am, f)$ as in model (3), then $F^*$ in the solution (4) is the convex conjugate of the function $F(Am, f)$. Followed by the PDHG-net-Ⅰ, a learned operator $\Gamma$ is also used to replace $prox_\sigma[F^*]$, and the new algorithm, PDHG-net-Ⅱ, can be written as

$$\begin{cases} d_{n+1} = \Gamma(d_n + \sigma A\bar{m}_n, f) \\ m_{n+1} = \Lambda(m_n - \tau A^T d_{n+1}) \\ \bar{m}_{n+1} = m_{n+1} + \theta(m_{n+1} - m_n) \end{cases} \quad (8)$$

The primal proximal $\Lambda$, dual proximal $\Gamma$, parameters $\sigma$, $\tau$ and $\theta$ are all learned from training data. To improve the representation capacity of the network, the parameters of the network in each iteration are different, which makes the network a cascading architecture.

### *PDHG-net-III*

To utilize the learning ability of deep networks better and further improve the reconstruction quality based on PDHG-net-Ⅱ, the explicitly enforced updating structures $d_n + \sigma A\bar{m}_n$, $m_n - \tau A^* d_{n+1}$ were broken and the combinations of the variables were freely learned by the network. Instead of the hard acceleration step $m_{n+1} + \theta(m_{n+1} - m_n)$, the network can be designed to freely learn in what point the forward operator should be evaluated. Thus, the algorithm, called PDHG-net-Ⅲ, is formulated as

$$\begin{cases} d_{n+1} = \Gamma(d_n, Am_n, f) \\ m_{n+1} = \Lambda(m_n, A^T d_{n+1}) \end{cases} \quad (9)$$

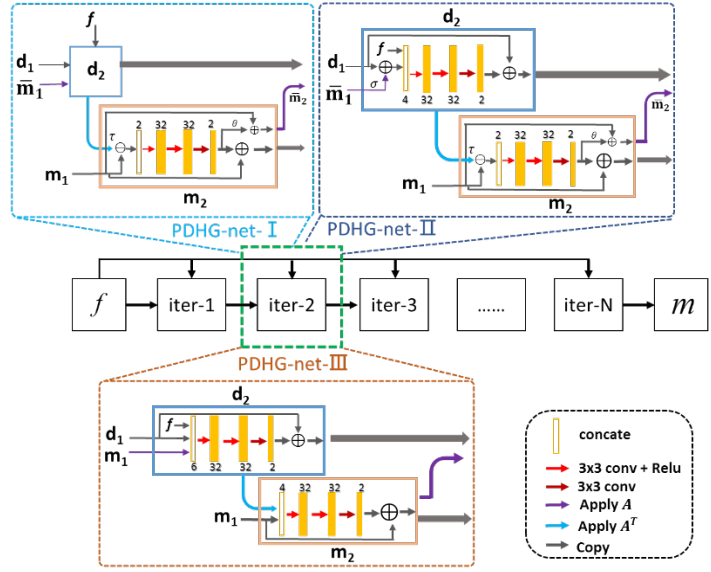

Fig 1. Architectures of the PDHG-nets.

② **the ADMM approach**

*ADMM-net-I*

ADMM-net was designed by unrolling the Alternating Direction Method of Multipliers (ADMM) algorithm [22]. The original network, denoted as basic-ADMM-CSNet, learns the regularization parameters in the ADMM algorithm [18]. It was then generalized in their follow-up work, denoted as Generic-ADMM-CSNet [23] The improved network learns the image transformation and regularizer in the regularization function. Here, the Generic-ADMM-CSnet (called ADMM-net-I hereafter) was used as the example.

In the context of ADMM-net-I, the regularization term in model (2) is written as

$$\lambda R(m) = \sum_{l=1}^{L} \lambda_l \|\psi_l m\|_p \tag{10}$$

where $\psi_l$ denotes a transformation matrix (e.g., discrete wavelet transform for a sparse representation), $\lambda_l$ is the regularization parameter.

Introducing a set of independent auxiliary variables $z = \{z_1, z_2, \cdots, z_L\}$ in the spatial domain, ADMM reconstructs the image by solving the following subproblems:

$$\begin{cases} \underset{m}{\mathrm{argmin}} \frac{1}{2}\|Am - f\|_2^2 + \frac{\rho}{2}\|m + \beta - z\|_2^2 \\ \underset{z}{\mathrm{argmin}} \sum_{l=1}^{L} \lambda_l \|\psi_l z\|_p + \frac{\rho}{2}\|m + \beta - z\|_2^2 \\ \underset{\beta}{\mathrm{argmin}} \sum_{l=1}^{L} \langle \beta, m - z \rangle \end{cases} \tag{11}$$

The solution is

$$\begin{cases} M^{(n)} : m^{(n)} = \frac{[A^T f + \rho(z^{(n-1)} - \beta^{(n-1)})]}{(A^T A + \rho I)} \\ Z^{(n)} : z^{(n)} = \mu_1 z^{(n,k-1)} + \mu_2 (m^{(n)} + \beta^{(n-1)}) \\ \qquad - \sum_{l=1}^{L} \tilde{\lambda}_l \psi_l^T H(\psi_l z^{(n,k-1)}) \\ P^{(n)} : \beta^{(n)} = \beta^{(n-1)} + \tilde{\eta}(m^{(n)} - z^{(n)}) \end{cases} \tag{12}$$

where $H(\cdot)$ refers to a non-linear operation corresponding to the gradient of the regularizer $\|\cdot\|_p$. The second subproblem is solved approximately by directly employing the gradient descent algorithm, in which the iteration is indexed by $k$. In Eq. (12), all parameters $(\rho, \mu_1, \mu_2, \lambda_l, \tilde{\eta})$ are learnable and the transformations $(\psi_l, \psi_l^T)$, which are implemented linearly by convolving with kernels, as well as the nonlinear operator $H$, which is implemented by non-linear function ReLU in our experiments, are also learnable.

### ADMM-net-II

In Eq. (11), the sparse regularization term is learnable whereas the data consistency is measured by the L2 norm of the difference between the estimated and the acquired data at the sampled locations. It would be more accurate if the consistency in k-space is learned by the network from the training data as in model (3). With the ADMM algorithm, we propose to solve the model (3) with the regularization term (10) as follows:

$$\begin{cases} \underset{m}{\operatorname{argmin}} F(Am, f) + \frac{\rho}{2}\|m + \beta - z\|_2^2 \\ \underset{z}{\operatorname{argmin}} \sum_{l=1}^L \lambda_l \|\psi_l z\|_p + \frac{\rho}{2}\|m + \beta - z\|_2^2 \\ \underset{\beta}{\operatorname{argmin}} \sum_{l=1}^L \langle \beta, m - z \rangle \end{cases} \quad (13)$$

Inspired by the solution to the second subproblem in (11), the solution to (13) is as follows:

$$\begin{cases} M^{(n)}: m^{(n)} = \gamma_1 m^{(n,k-1)} + \gamma_2 \left( z^{(n)} - \beta^{(n-1)} \right) - A^T \Gamma(Am^{(n,k-1)}, f) \\ Z^{(n)}: z^{(n)} = \mu_1 z^{(n,k-1)} + \mu_2 \left( m^{(n)} + \beta^{(n-1)} \right) - \sum_{l=1}^L \tilde{\lambda}_l \psi_l^T H\left( \psi_l z^{(n,k-1)} \right) \\ P^{(n)}: \beta^{(n)} = \beta^{(n-1)} + \tilde{\eta}\left( m^{(n)} - z^{(n)} \right) \end{cases} \quad (14)$$

where $\Gamma(Am, f)$, which is accomplished by the neural network, refers to the function corresponding to the deviation of data consistency $F(Am, f)$.

### ADMM-net-III

In ADMM-net-II, sparse prior is imposed as the prior information, with the powerful learning ability of deep networks, the prior besides sparsity could be learned from the training data, thus the reconstruction model, as listed in (3), becomes more general.

The second subproblem in (13) then becomes

$$\underset{z}{\operatorname{argmin}} R(z) + \frac{\rho}{2}\|m + \beta - z\|_2^2 \quad (15)$$

Noticed that (15) is the same as (5), which is the proximal operator that can be replaced by a learned network. Followed the strategy of developing PDHG-net-III, we propose to freely learn the combinations of variables in the first subproblem of (13) by the network. So the solution iterations can be written as

$$\begin{cases} D^{(n)}: d^{(n)} = \Gamma(Am^{(n-1)}, f) \\ M^{(n)}: m^{(n)} = \Pi(m^{(n-1)}, z^{(n-1)} - \beta^{(n-1)}, A^T d^{(n)}) \\ Z^{(n)}: z^{(n)} = \Lambda(m^{(n)} + \beta^{(n-1)}) \\ P^{(n)}: \beta^{(n)} = \beta^{(n-1)} + \tilde{\eta}(m^{(n)} - z^{(n)}) \end{cases} \quad (16)$$

The operator $\Gamma$, $\Pi$, $\Lambda$ and the parameter $\tilde{\eta}$ are all learned by the network.

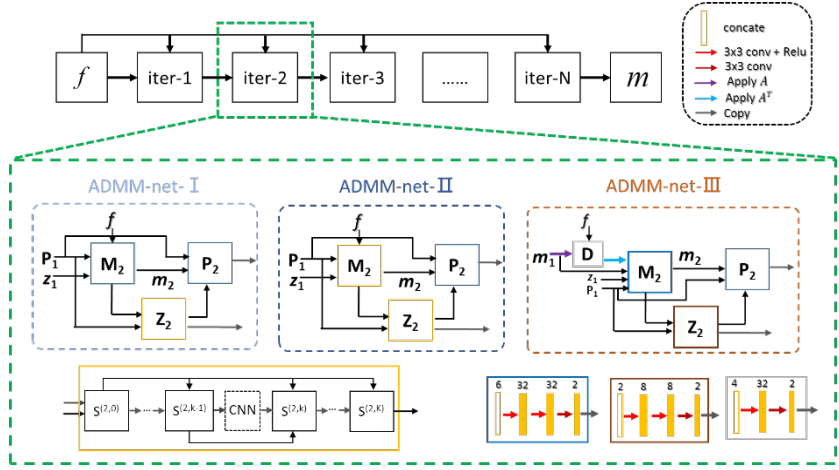

Fig 2. Architectures of the ADMM-nets.

③ **the ISTA approach**

*ISTA-net-I*

Iterative shrinkage-thresholding algorithm (ISTA) is a popular first-order method to solve linear inverse problem [24]. For the CS reconstruction problem (2) with the regularization $\|\Psi m\|_1$, ISTA reconstructs the images with following iterations:

$$\begin{cases} r^{(n+1)} = m^{(n)} - \rho A^T(Am^{(n)} - f) \\ m^{(n+1)} = \underset{m}{\mathrm{argmin}} \frac{1}{2} \|m - r^{(n+1)}\|_2^2 + \tau \|\Psi m\|_1 \end{cases} \quad (18)$$

In ISTA-net-I, a general nonlinear transform function $G$ is adopted to sparsify the images, whose parameters are learnable. Therefore, iteration (18) becomes

$$\begin{cases} r^{(n+1)} = m^{(n)} - \rho A^T(Am^{(n)} - f) \\ m^{(n+1)} = \underset{m}{\mathrm{argmin}} \frac{1}{2} \|G(m) - G(r^{(n+1)})\|_2^2 + \tau \|G(m)\|_1 \end{cases} \quad (19)$$

The solution is

$$\begin{cases} r^{(n+1)} = m^{(n)} - \rho A^T(Am^{(n)} - f) \\ m^{(n+1)} = \tilde{G}\left(soft(G(r^{(n+1)}), \theta)\right) \end{cases} \quad (20)$$

the step size $\rho$, shrinkage threshold $\theta$, forward transform $G$ and the backward transform $\tilde{G}$ are the learnable parameters in ISTA-net-I.

*ISTA-net-II*

Based on the ISTA-net-I, we further relax the data consistency term $\|Am - f\|_2^2$ as $F(Am, f)$, then the deviation of $F(Am, f)$ can be written as $A^T \Gamma(Am, f)$, in which the function $\Gamma(Am, f)$ can be accomplished by the deep network. Therefore, the iteration of ISTA-net-II becomes

$$\begin{cases} d^{(n+1)} = \Gamma(Am^{(n)}, f) \\ r^{(n+1)} = m^{(n)} - \rho A^T d^{(n+1)} \\ m^{(n+1)} = \tilde{G}\left(soft(G(r^{(n+1)}), \theta)\right) \end{cases} \quad (21)$$

*ISTA-net-III*

In ISTA, the residual $r^{(n+1)}$ is imposed by the difference between the current solution and the deviation of data consistency term, we then propose to use the network to learn the combination of these variables. Thus the solution is as follows:

$$\begin{cases} d^{(n+1)} = \Gamma(Am^{(n)}, f) \\ r^{(n+1)} = \Lambda(m^{(n)}, A^T d^{(n+1)}) \\ m^{(n+1)} = \tilde{G}\left(soft(G(r^{(n+1)}), \theta)\right) \end{cases} \quad (22)$$

The operators $\Gamma$, $\Lambda$, $G$ and $\tilde{G}$ are all realized by networks.

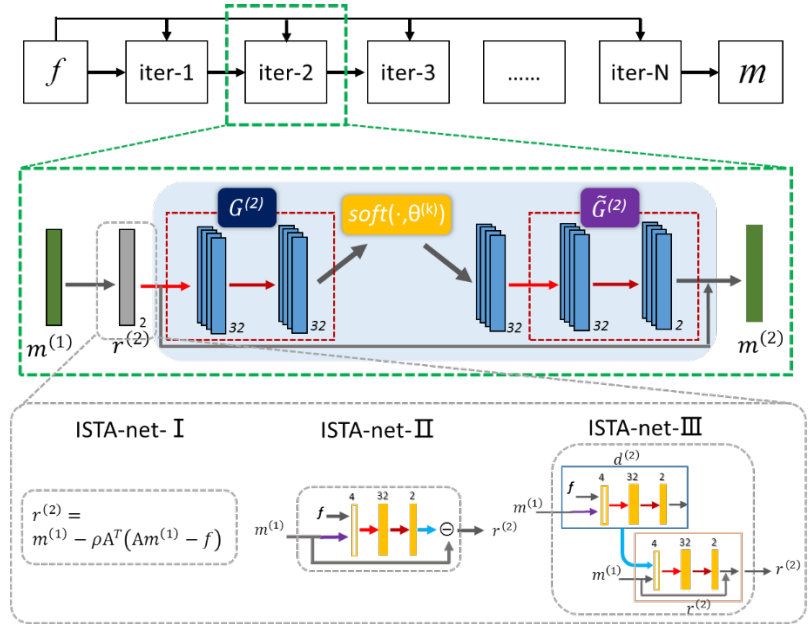

Fig 3. Architectures of the ISTA-nets.

**2.3 Two hypotheses**

We have one hypnosis on these three levels of relaxation, and another hypnosis on the difference of the learning ability of the networks which root from these three algorithms.

***Rule1: Given an established CS model consisting of data consistency and regularization***

*terms, the more the items are relaxed, the stronger the learning ability of the unrolling network has, which could produce better reconstructions.*

This hypnosis can be theoretically explained by the concept of VC dimension which is fundamental in machine learning. Fig. 4 shows the illustration on the error curves of out-of-sample (testing error), model complexity and in-sample (training error) with the increasing of VC dimension.

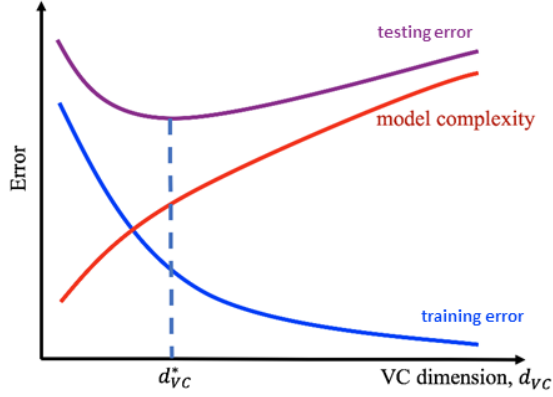

Fig 4. VC dimension.

In our work, the model complexity corresponds to the number of network parameters, which increases with the network becoming more complicated from state I to III. Thus, the in-sample error decreases and the out-of-sample error also decreases and the network III may approach the optimal VC dimension.

*Rule2: Given an established CS model consisting of data consistency and regularization terms, when we derive the iterative steps for a specific algorithm, the less the approximation is used, the stronger the learning ability of the unrolling network has, which could produce better reconstructions.*

In the derivation of PDHG, the first-order (Taylor) approximation was applied in both primal and dual space (Eq.4). PDHG is a variant of the projected gradient method in some cases. In the derivation of ISTA (Eq.18), the first-order Taylor approximation was applied in the data-consistency term. While in ADMM, no approximation (constraint) is enforced in the derivation procedure (Eq.11).

## 3. Method

**3.1 Network architecture**

The network architectures of each algorithm mentioned above are derived from the data flow of the original optimization methods. The convolutions in our paper are all 3×3 pixels in size, and the nonlinear operator is chosen to be Rectified Linear Unites (ReLU). As the data consistency is realized by the neural network, algorithm of state II is deeper than state I for each approach. The output of each CNN block has two channels representing the real and imaginary parts in the task of MR reconstruction as MR data is complex-valued.

The entire architecture of primal dual networks and one iteration block of the three states of PDHG-net are illustrated in Fig. 1. The primal and dual iterations have the same architecture with three convolutional layers in each block of PDHG-net-II and PDHG-net-III. To train the network more easily, residual network was used. For PDHG-net-II, the number of channels is 2-32-32-2 in each primal update and 4-32-32-2 in each dual update, whereas for PDHG-net-III, the number of channels is 4-32-32-2 in each primal update and 6-32-32-2 in each dual update. We set the number of iterations to be 10 in all three networks.

The data flow graphs of the ADMM networks for one iteration are demonstrated in Fig. 2. The graph nodes of ADMM-net-I, ADMM-net-II and ADMM-net-III correspond to the operations in Eq. (11), Eq. (14) and Eq. (17), respectively. For ADMM-net-I, the operation $\widetilde{\lambda}_l D_l^T H(D_l z^{(n,k-1)})$ in $Z^{(n)}$ is accomplished by a CNN module, which comprises of two convolution layers and a non-linear activation layer with 8 filters. The reconstruction module $M^{(n)}$ in ADMM-net-II is as the same structure to $Z^{(n)}$, but with 32 filters. The number of iterations for the ADMM-nets are all 15 as suggested in the original ADMM-net-I work [23].

The second iteration of ISTA-nets is illustrated in Fig. 3. To give a better reconstruction, the image was sparsified in the residual domain, which leads to a sparser representation and make it a residual network to facilitate the training. The number of filters we used for ISTA-net-I is 32. In ISTA-net-II, $\Gamma(Am^{(n)}, f)$ in Eq. (21) is modeled as two convolution operators and one ReLU operator. 10 iterations were used here for ISTA-nets.

**3.2 Evaluation on raw MR scanner data**

*Training data for MRI*

We demonstrate our model-based deep learning image reconstruction framework using MRI as a model system, but we emphasize that our framework is applicable to image reconstruction problems across a broad range of image modalities where traditional CS-based methods are popular. Overall 2100 fully sampled multi-contrast data from 10 subjects with a 3T scanner (MAGNETOM Trio, SIEMENS AG, Erlangen, Germany) were collected and informed consent was obtained from the imaging object in compliance with the IRB policy. The fully sampled data was acquired by a 12-channel head coil with matrix size of 256×256 and adaptively combined to single-channel data [25] and then retrospectively undersampled using Poisson disk sampling mask.

*Testing data for MRI*

We tested the proposed methods on 398 human brain 2D slices from 3D data acquired from SIEMENS 3T scanner with 32-channel head coil and 3D T1-weighted SPACE sequence, TE/TR=8.4/1000ms, FOV=25.6×25.6×25.6cm³. The data was fully acquired and then manually combined to single-channel data and down-sampled for reconstruction.

The proposed methods also have been tested on the fully sampled data from other commercial 3T scanners (GE Healthcare, Waukesha, WI; United Imaging Healthcare, Shanghai, China).

Multi-channel T2-weighted data were from the reference [17], which were acquired using a 3D T2 CUBE sequence with a 12-channel head coil. The matrix size is 256×232×208 with 1mm isotropic resolution. Fully sampled brain data from four volunteers were used for training with in total 360 slices, and the data with in total 164 slices from one volunteer was retrospectively undersampled using 6X Poisson disk sampling mask for testing. The coil sensitivity maps were pre-estimated using ESPIRiT [26]. Please note, the sampling masks from each training and testing slice are different.

### 3.3 Evaluation on simulated CT data

To validate the feasibility of the proposed network on other imaging modalities, we used data from an authorized clinical low-dose CT dataset, which was made for the 2016 NIH AAPM-Mayo Clinic Low-Dose CT Grand Challenge. Normal-dose abdominal CT images (image size was downsampled to 128×128 for computation convenience) of 6 patients were taken as labels, and sparse-view sinograms were simulated by forward projection of labels with specific fan-beam geometry. The distance from X-ray source to detector is 1200mm and the distance from X-ray source to rotation center is 1000mm. The linear-array detector has 300 elements with element size of 0.5mm×0.5mm. Each rotation includes 90 projection views evenly spanned on a 360° circular orbit. We used 1600 sinogram-label image pairs of 5 patients for training and 360 pairs of another patient for testing. The number of iterations was set to 15 for ADMM-nets.

### 3.4 Evaluation on simulated PET data

Clinical PET images from 10 patients scanned on a GE Discovery PET/CT 690 machine at a single institution were used. Data from 6 patients were used for training, from 2 patients for validation, and from the remaining 2 patients for testing. Each patient has 310 Dicom format images. All of the images were then forward-projected to generate corresponding noise-free sinograms, each having a size of 180 × 273 pixels. Then, we normalized the sum of the sinograms count to 5e6 (about 10% percent of the normal dose) and introduce Poisson noise. To simulate the scatter and random fraction, a uniform background of 20% total true coincidences was added. After these steps, one sinogram and one reference PET image were paired to generate the training, validation and testing datasets.

We compare ADMM-nets to other methods, including MLEM and FBP. In the comparison of results, the number of iterations of the MLEM method was set to 60. And the FBP filter is a Ram-Lak filter multiplied by a Hamming window.

### 3.5 Evaluation metrics

In network training, the mean square error (MSE) is chosen as the loss function. Given pairs of training data, the loss between the network output and ground truth is defined as

$$L_{MSE}(\Theta) = \frac{1}{N}\sum_{i=1}^{N}\|\hat{m}(\Theta,f) - m^{ref}\|_2^2 \qquad (23)$$

where $\hat{m}(\Theta,f)$ is the network output based on network parameter $\Theta$ and sampled data $f$, $m^{ref}$ is the corresponding ground truth. Specifically, because of the symmetric constraint in the ISTA-nets, the loss function in ISTA-nets is designed as follows:

$$L(\Theta) = L_{MSE}(\Theta) + \gamma L_{constraint}(\Theta) \qquad (24)$$

where $L_{MSE}(\Theta)$ is defined as (23) and $L_{constraint}(\Theta)$ is imposed to satisfy the symmetric constraint, which is defined as

$$L_{constraint}(\Theta) = \frac{1}{N}\sum_{i=1}^{N}\sum_{k=1}^{N_p} \left\| \tilde{G}^{(k)}\left(G^{(k)}(\hat{m})\right) - \hat{m} \right\|_2^2 \tag{25}$$

where $N_p$ is the total number of iterations. In our experiments, $N_p = 10, \gamma = 0.01$.

Several similarity metrics, including MSE, structural similarity (SSIM) and peak signal-to-noise ratio (PSNR), were used to compare the reconstruction results of different methods with the reference image from fully sampled data.

We trained the networks by minimizing the loss function defined above using the ADAM optimizer in TensorFlow. And the trainings were performed on an Ubuntu 16.04 LTS (64-bit) operating system equipped with a Tesla TITAN Xp Graphics Processing Unit (GPU, 12GB memory) with CUDA and CUDNN support.

## 4. Result

As the constraints in the specific model (1) are gradually relaxed, the quality of the reconstruction gets better and better, which is shown in Fig. 5. From the state I to the state III, the reconstruction model becomes more general, which is beneficial for the network to learn the property of the training data, including but not limited to sparsity, thus improves the image quality. Nevertheless, the required training set is expected to increase for achieving the optimal performance because of the increasing number of parameters to be learned.

Fig 6 shows the visual comparisons with the acceleration factor of 6. The zoom-in images of the enclosed part and the corresponding error maps as well as the quantitative metrics are also provided. With the constraints relaxed, the fine details are able to be recovered, as the prior information is learned from the training data, which is more suitable than the hand-designed prior for the undersampled data. Fig 7 shows the visual comparisons on the multi-channel data with the acceleration factor of 6. With the constraints relaxed, the quality of reconstructed images become better and better, which is also verified by the quantitative measurements.

Fig 8 shows the visual comparisons on the simulated CT data with 90 views. We can see that FBP reconstruction exhibits obvious striking artifacts, while ADMM-net-I reconstruction only exhibits few artifacts. With the constraints relaxed, the reconstructions of ADMM-net-II and ADMM-net-III have no visual artifacts and the latter preserves more details.

Fig 9 shows the visual comparisons on the simulated PET data with 10% dose. ADMM-net-III exhibits the best coincidence with the reference in terms of artifacts removing and detail preserving.

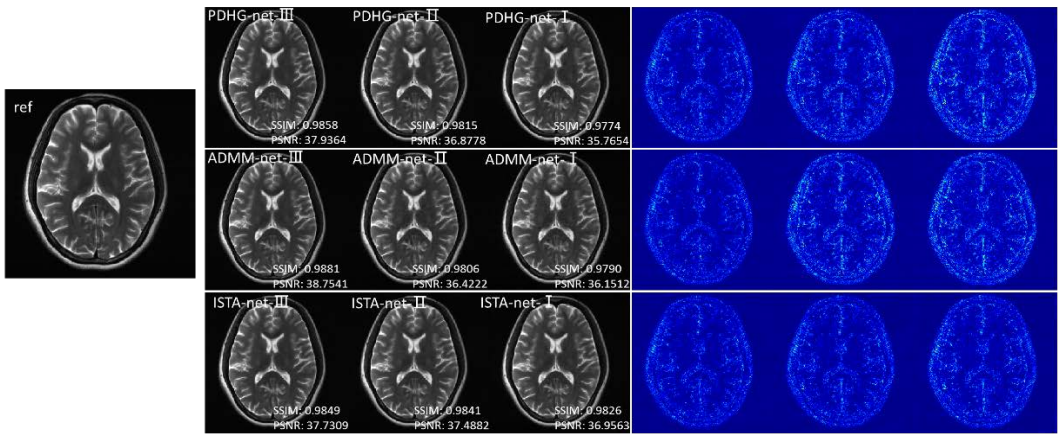

Fig 5. Reconstruction results and the corresponding error maps when gradually relaxing constraints with respective to the three algorithms. A 10x Poisson disk sampling mask was used on an axial data from the UIH scanner.

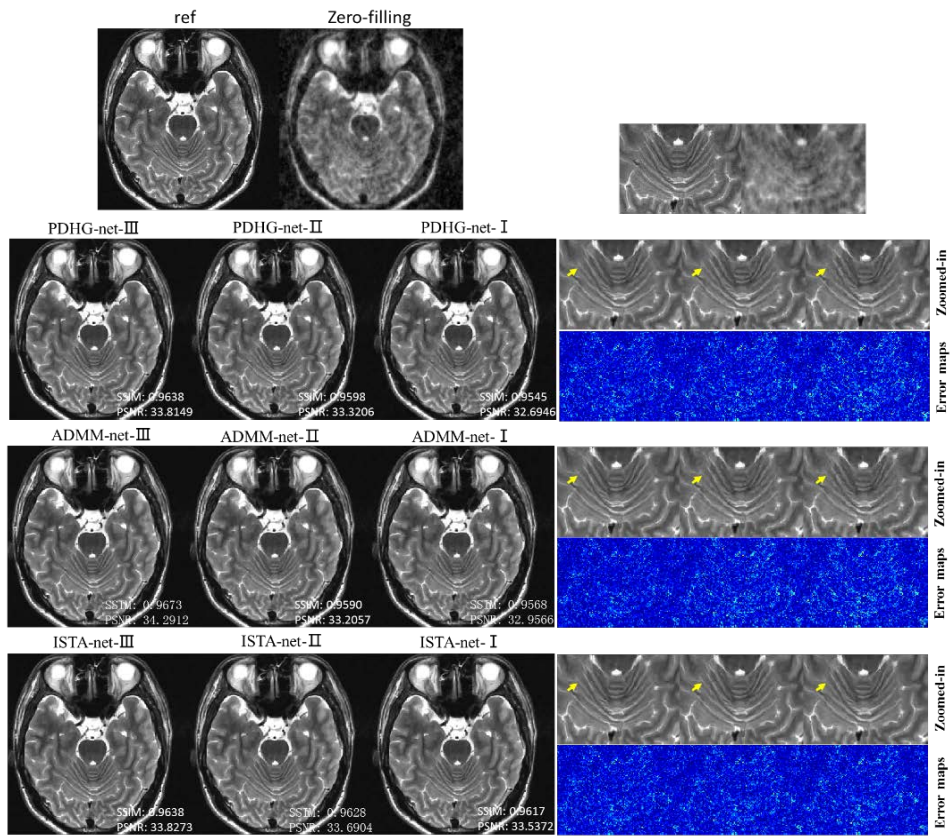

Fig 6. Reconstruction results with a 6x Poisson disk sampling on an axial data from the Siemens scanner. The zoomed in images and the corresponding error maps are provided on the right, respectively.

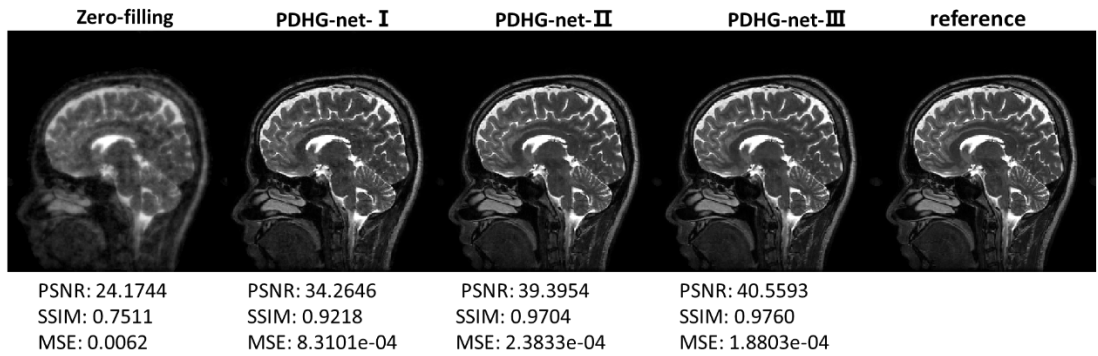

Fig 7. Reconstruction results with a 6x Poisson disk sampling on a 12-channel sagittal data from the GE scanner.

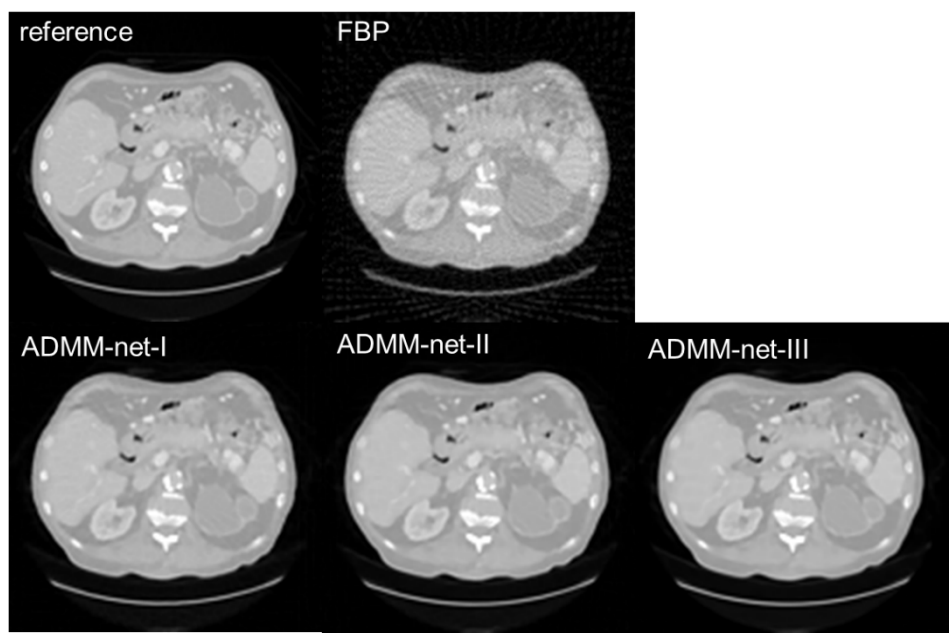

Fig 8. Reconstruction of CT simulated data with 90 views.

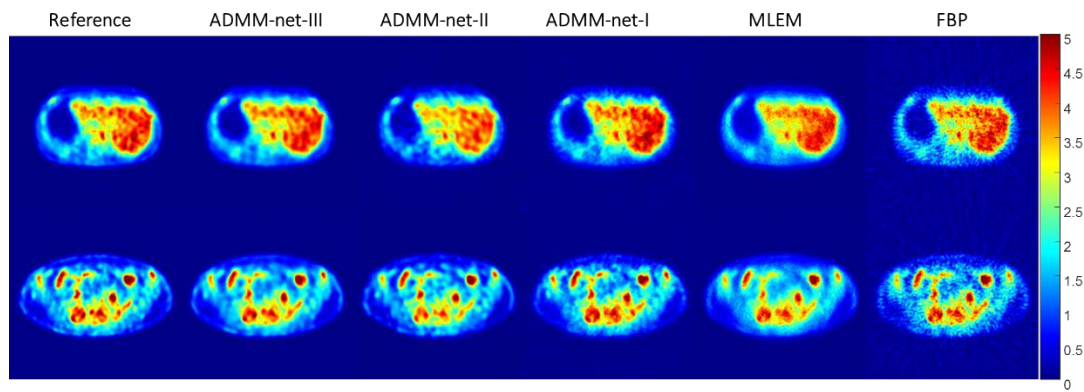

Fig 9. Reconstruction of PET simulated data with 10% percent dose.

# 5. Discussion

**5.1 Training set**

The superior performance achieved by deep learning methods heavily relies on the large training dataset. For medical image reconstruction, data-driven methods usually demand more training data than model-driven methods. Since the solution space is huge without model, data-driven methods search the mapping between the training pairs on the large dataset to prevent the solution from local minimum, which is the issue of overfitting. Whereas for model-driven methods, the solution space is narrowed by the reconstruction model, and the solution searching is guided by the iterative inference algorithm, thus the training data for model-driven methods is reduced.

In this paper, we gradually relax the requirements on a specific reconstruction model (e.g. CS model). With more uncertainty in the model, that is increasing the degree of freedom of the network, the size of training data is expected to be increased for achieving the optimal performance. As shown in Fig. 10, with little training data, the method with more relaxed constraints (state III) has comparable performance or even worse than the methods with less relaxed constraints (states I and II). As the training data increases, the performance of state III changes more significantly than others. With sufficient training data, state III gives the best reconstruction.

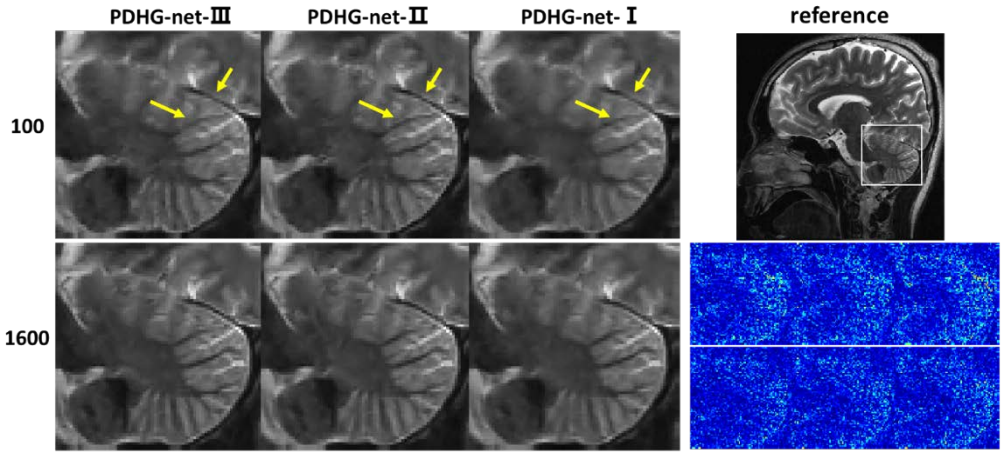

Fig 10. The reconstructed zoom-in images of the enclosed part with a 6x Poisson disk sampling on a sagittal data from the Siemens scanner. The results with a small training size are located in the first row on the left and the second row shows the results with more training data. The corresponding error maps of the enclosed part are also provided on the right bottom.

**5.2 Stability and reconstruction accuracy**

The stability of the network is an important factor to be considered in practice. Given that the network is trained with a specific sampling pattern, the question is how robust the trained network with respect to the changes of acceleration factor is. In our experiments, all the networks are trained with the 6-fold Poisson disk sampling mask. We have tested the performance of the networks with other sampling ratios, shown in the Fig. 11. The results show the flexibility of the networks for changing the amount of acquired data. The quality of reconstruction deteriorates with less samples for each state, whereas for the fixed acceleration, the improvement of quality induced by relaxing can be observed. Moreover, the average reconstruction time with R=6 are listed in Table 2.

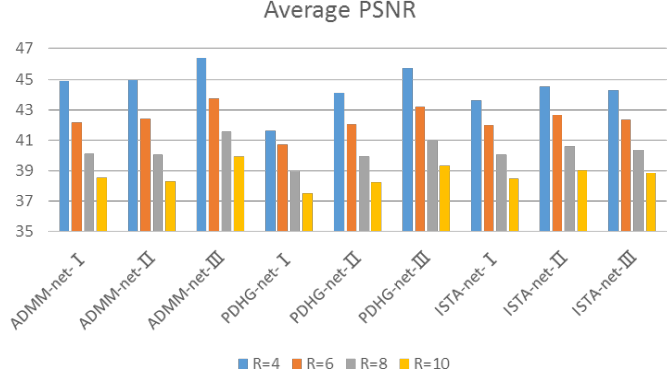

Fig 11. The average PSNR performance of the all networks with different acceleration factors.

Table 2. Average reconstruction time comparison between the methods.

|  | I | II | III |
|---|---|---|---|
| PDHG-net | 0.03517s | 0.03904s | 0.03685s |
| ADMM-net | 0.12252s | 0.15485s | 0.14938s |
| ISTA-net | 0.03511s | 0.03741s | 0.03850s |

CS-based approaches suffer from detail loss at large acceleration factors, several attempts have been made to address this issue [27, 28]. Although deep learning reconstruction is superior to CS, it still exhibits feature loss as CS does due to the l2-norm loss function, which is employed by most DL approaches. In this paper, the results are with higher quality in terms of various quantitative values compared with the original version, but still exhibit over smoothing at high frequency region. Figure 12 shows the error spectrum plots (ESP) [29] with Fourier radial of the three algorithms with the 6x sagittal MR data from GE scanner. From the ESP, the relative error increases as the Fourier radial increases, which indicates that the high frequency part of the image has higher error than low frequency part. In the specific algorithm, the network with state III has the lowest error. This is consistent with the visual comparison.

In our experiments, the sub-network architectures and the parameters of the same module in all three states are the same to give a fair comparison. Another consideration is whether the improvement from state I to state II is induced by the learned data fidelity rather than the deeper network. We have added convolutional layers to the primal modules of PDHG-net-I to make it have the same number of parameters as PDHG-net-II, the new network was denoted as PDHG-net-I*. The average PSNR performance and the visual comparison on the 6x testing data are shown in Fig. 13 and Fig. 14, respectively. PDHG-net-I* performs better than PDHG-net-I because of the deeper network with more learning parameters, whereas it is not comparable to PDHG-net-II, the reason is that the learned data fidelity is more suitable than the predefined l2-norm for the data.

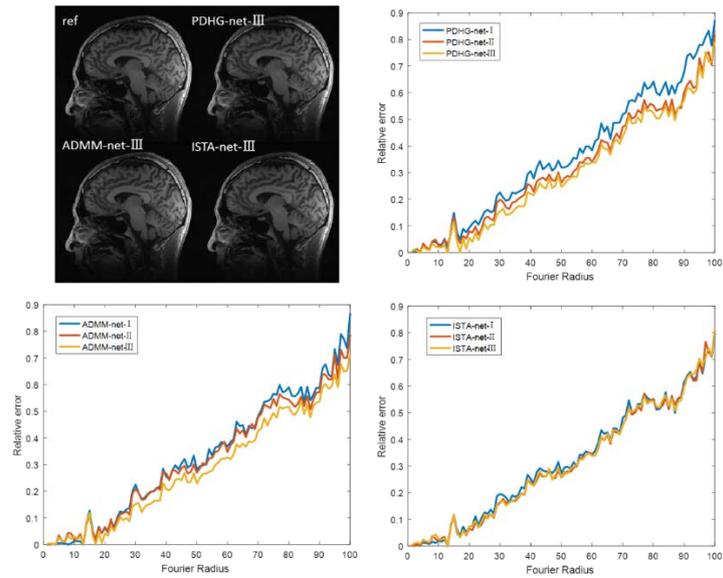

Fig 12. Error Spectrum Plots of the three types of algorithms.

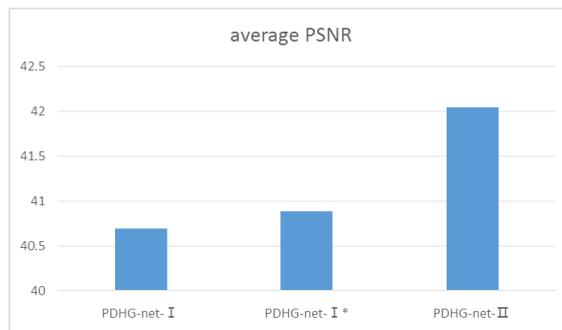

Fig 13. The effectiveness of learned data fidelity. PDHG-net-I* has the same structure as PDHG-net-I but the same number of parameters as PDHG-net-II.

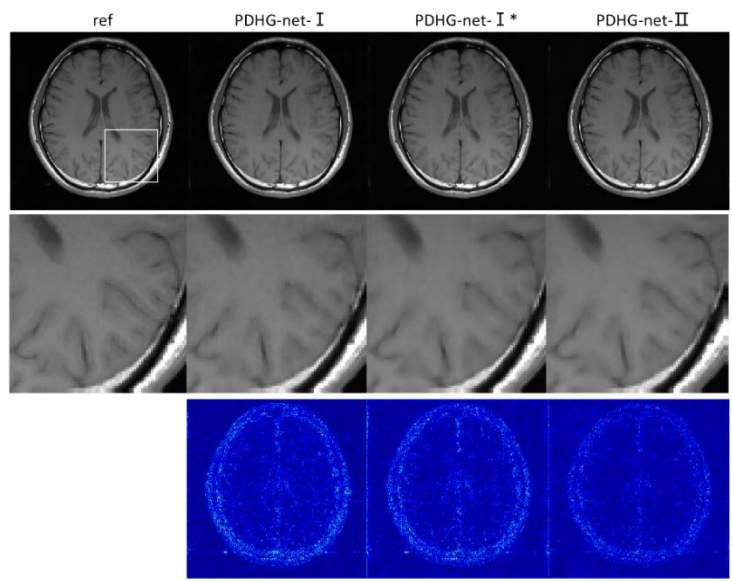

Fig 14. Reconstruction results with a 6x Poisson disk sampling on an axial data from the GE scanner.

**5.3 Network interpretation**

For linear inverse problem, signal priors are often used to regularize the inverse problem. Traditionally, these priors are hand-designed based on empirical observations of images such as transform sparsity. Though these priors can be used in many inverse problems related to images and usually have efficient solvers, they may have poor performance on a specific problem. For an individual problem like image reconstruction, hand-crafted priors are too general to constraint the solution set, learning can fit the data property better if the testing data is in the same type of training data. That's why learning-based methods perform better than traditional methods.

For the model-driven deep learning methods, the sub-network at each iteration is designed with the data flow graphs, which are derived from the iterative procedure of the optimization algorithm. Although the networks of state Ⅲ have little constraints in the model and inference, they indeed root from the corresponding mathematical algorithms with the module of network being the same as the data flow graphs of the inferences. The network, especially the convolutional neural network used here, is adopted to parameterize the solution space due to the strong representative ability, in order to make the inference more suitable for the specific applications.

**5.4 Open issues in DL reconstruction**

Although the optimization algorithm induced networks can learn the parameters in the algorithm and achieve superior performance, developing network architectures is still a time-consuming and error prone process, since all the hyper-parameters of the network such as learning rate and number of feature maps are decided empirically. In our experiments, the hyper-parameters of the networks in one algorithm chain are the same for fair comparison. For example, the learning rate, number of iterations, number of layers and the feature maps of the same module are set to be the same.

Recently, network architecture search (NAS) is considered as a feasible and promising tool to automatically search for the optimal network architecture [30]. Several novel NAS methods, such as reinforcement learning and evolutionary computation, have been developed and applied in a variety of deep learning tasks [31-33]. It also can be applied in medical image reconstruction in the future.

Besides the hyper-parameters, loss function is another important factor need to be considered in network design. Loss function guides the learning direction, and different loss functions may result in different reconstructions. The MSE we used here is prone to fail to recover tiny structures in the images, as MSE is an average indicator of the whole image, more advanced loss functions related to the integrant of interest could be applied [34].

# 6. Conclusion

In this paper, we developed an effective framework to integrate classical inference and deep neural network to maximize the potential of deep learning and model-based reconstruction for medical imaging. The experimental results on simulated and in vivo data demonstrate the effectiveness of the proposed framework. This work attempts to provide a guideline on how to improve the image quality with deep learning, based on the traditional iterative reconstruction algorithms. More techniques and properties of the unification of deep learning and traditional reconstruction methods should be investigated in the future.